\documentclass[sigconf]{acmart}

\renewcommand\footnotetextcopyrightpermission[1]{}
\setcopyright{none}

\acmConference[]{}{}{}
\acmBooktitle{}
\acmDOI{}
\acmISBN{}

\AtBeginDocument{
  
}

\usepackage{amsmath}
\usepackage{graphicx}
\usepackage{booktabs}
\usepackage{amsthm}
\usepackage{multirow}

\newtheorem{problem}{Problem}
\renewcommand\footnotetextcopyrightpermission[1]{}
\acmConference[]{}{}{}
\acmBooktitle{}
\acmPrice{}
\acmDOI{}
\acmISBN{}

\title{Optimizing Coverage and Difficulty in Reinforcement Learning for Quiz Composition}

\author{Ricardo Pedro Querido Andrade Silva}
\affiliation{
  \institution{CNRS, Univ. Grenoble Alpes}
  \city{Grenoble}
  \country{France}
}

\author{Nassim Bouarour}
\affiliation{
  \institution{CNRS, Univ. Grenoble Alpes}
  \city{Grenoble}
  \country{France}
}

\author{Dina Fettache}
\affiliation{
  \institution{CNRS, Univ. Grenoble Alpes}
  \city{Grenoble}
  \country{France}
}

\author{Sarab Boussouar}
\affiliation{
  \institution{CNRS, Univ. Grenoble Alpes}
  \city{Grenoble}
  \country{France}
}

\author{Noha Ibrahim}
\affiliation{
  \institution{Universit\'e Grenoble Alpes}
  \city{Grenoble}
  \country{France}
}

\author{Sihem Amer-Yahia}
\affiliation{
  \institution{CNRS, Univ. Grenoble Alpes}
  \city{Grenoble}
  \country{France}
}

\begin{document}
\fancyhead{}
\renewcommand{\headrulewidth}{0pt}
\begin{abstract}
Quiz design is a tedious process that teachers undertake to evaluate the acquisition of knowledge by students.  Our goal in this paper is to automate quiz composition from a set of multiple choice questions (MCQs). We formalize a generic sequential decision-making problem with the goal of training an agent to compose a quiz that meets the desired topic coverage and difficulty levels. We investigate DQN, SARSA and A2C/A3C, three reinforcement learning solutions to solve our problem. We run extensive experiments on synthetic and real datasets that study the ability of RL to land on the best quiz. Our results reveal subtle differences in agent behavior and in transfer learning with different data distributions and teacher goals. This was supported by our user study, paving the way for automating various teachers' pedagogical goals.
\end{abstract}
\maketitle

\section{Introduction}
Teachers spend considerable time crafting quizzes to evaluate their students' knowledge acquisition~\cite{Olney2023,Meissner2024,Maity2024}. They usually follow a stepwise decision-making process starting from a mix of previously used quizzes and newly designed Multiple-Choice Questions (MCQs). At each step, they seek to ensure that the MCQs forming a quiz cover desired topics and difficulty level distributions. In this paper, our aim is to help teachers automate quiz composition.\\

\begin{figure*}[!htbp]
\includegraphics[width=\textwidth]{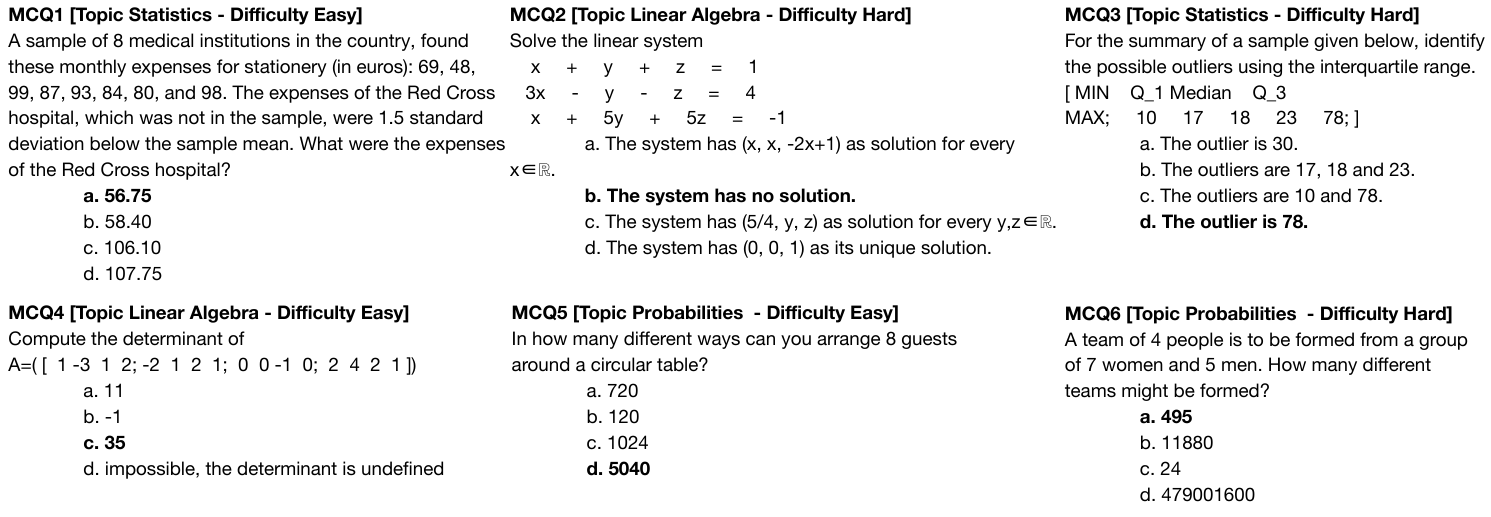}
    \caption{\small MCQs used to generate quizzes.}
    \label{fig:MCQs}
\end{figure*}

\noindent{\bf Context and motivating example.} Quiz creation has recently moved from manual editing to smarter, learning-based systems. Teachers used to spend a lot of time adjusting questions one by one to create a good quiz. Key concerns like topic coverage, difficulty balance, and relevance to the material are still essential for building high-quality quizzes. Consider the six Multi-Choice Questions provided in Figure~\ref{fig:MCQs} where the topic, difficulty level, and correct solution of each MCQ are highlighted in bold.  A teacher seeking to build a 3-MCQ quiz to test students on two topics, e.g. Statistics and Probabilities, would choose MCQs with varying difficulties, e.g., MCQs 1, 5, and 6 (this answer is not unique). A teacher who aims to vary topics and maintain the same difficulties would choose MCQs 1, 4 and 5, that constitute the only solution in this case. In practice, teachers proceed either by creating MCQs from scratch, or by replacing some MCQs by others until they reach a satisfactory quiz. 

To reduce human effort in composing quizzes from large MCQ pools, early systems used rule-based methods with predefined templates and constraints (e.g., “include two geometry questions”), offering limited flexibility and generalization \cite{alsubait2016, Heilman2010}. Later work automated quiz generation using difficulty prediction via lexical and syntactic features \cite{Feng2025}, or clustering and topic modeling to group or diversify questions \cite{Kurdi2020}. However, these methods target single objectives—such as topic relevance or redundancy reduction—and do not balance multiple pedagogical criteria.

\noindent{\bf Challenges.}
Selecting $k$ MCQs from a large pool is difficult because the system must incrementally choose questions that best meet target topic coverage and difficulty. As demonstrated in recent work~\cite{Shang2023}, this sequential decision process naturally suits RL. However, how different RL methods handle varying data distributions and teacher goals remains unclear. Key challenges include designing the MDP—especially rewards that reflect teacher objectives and actions that mimic quiz-design operations—and determining how to train agents for reusable performance across future quiz-design tasks.

\noindent{\bf Contributions.}
We define {\sc QuizComp}, a constrained bi-objective optimization problem that takes MCQs and target topic and difficulty distributions and outputs a satisfying quiz. We design an MDP where each state is a quiz, transitions replace it with one that improves topic or difficulty (reflected in the reward), and actions operate via similarity or dissimilarity on each objective. We implement three RL solutions: DQN, SARSA, and A2C/A3C. DQN serves as the benchmark and has prior use in quiz composition. SARSA enables comparison within temporal difference methods and has shown multitask potential. A2C/A3C provide a structurally different RL approach, combining policy and value networks with potential GPU efficiency benefits.

\noindent{\bf Empirical validation.} Our experiments evaluate multiple RL algorithms across varied input samples and topic/difficulty distributions using synthetic data and the real {\sc Med} and {\sc Math} datasets. Results confirm that agents can mimic teacher behavior using only MCQ similarity and dissimilarity. All algorithms use all actions, converge quickly, and tend to favor small gains, showing a bias toward topic and difficulty similarity. This leads to local exploration of MCQ neighborhoods, with larger jumps taken only when similar options are exhausted, mirroring human quiz design. On real datasets, all methods find quizzes highly aligned with targets, with DQN performing best.

We further test transferability across target types, uniform  and biased, and datasets, finding strong cross-domain and cross-target transfer. Agents trained on one dataset transfer well to the other, and those trained on biased targets transfer effectively other targets. Our user study with 28 qualified participants resulted in higher satisfaction and lower effort, confirming the need for automating quiz generation.

\section{Problem and Solutions} \label{sec:model}


We consider a set of knowledge topics $T$ and a set $M$ of MCQs. We associate to each MCQ $mcq \in M$ a topic $t \in T$ and a categorical value $l$ that reflects its difficulty level. Following common practice, we consider the difficulty levels in the Bloom taxonomy~\cite{Scaria2024,Maity2024,Hwang2023}. Our framework accommodates a variable number of difficulty levels, which depend on the dataset. We define a quiz as a subset $Z \subseteq M$ of MCQs of a fixed size $k$. We associate with each quiz $Z$, a topic vector of a fixed size, where each entry $i$ is computed as the proportion of MCQs in $Z$ with topic $t_i \in T$ (recall that each MCQ has a single topic). We also associate to $Z$ a difficulty vector of fixed size, where each entry is computed as the proportion of MCQs in $Z$ that are associated with a given difficulty level in the taxonomy.

A teacher wishing to compose a quiz has in mind  topics and difficulty levels to cover. Those are expressed as two target vectors: $T_C$, a distribution of proportions of MCQs in the quiz with desired topics, and $T_D$, a distribution of proportions of MCQs in the quiz of desired difficulty levels. For example, $<0, 0, 0, 0, 0, 0.5, 0, 0, 0.5, 0>$ is a biased topic vector where half the MCQs cover one topic and the other half another, and $<0.2, 0.2, 0.2, 0.2, 0.2>$ represents a uniform vector of difficulties. 

We define $\mathit{topicMatch(Z, T_C)}$ and $\mathit{diffMatch(Z, T_D)}$, two functions that reflect to what extent a quiz $Z$ reflects the desired distributions. Our formalization is agnostic to how these functions are defined. In our implementation, we use Cosine similarity.

\begin{problem}[The \textsc{QuizComp} Problem]
\label{problem:QuizComp}
Given a set $M$ of MCQs, two target vectors $T_C$ and $T_D$, and an integer $k$, our goal is to compose a quiz $Z \subseteq M$ of $k$ MCQs s.t.:
\begin{center}
\begin{flalign}
  \begin{aligned}
        &\ \texttt{argmax}_{Z \subseteq M} \ 
        \mathit{topicMatch} (Z, T_C) \\
        &\ \texttt{argmax}_{Z \subseteq M} \ \mathit{diffMatch} (Z, T_D)\\
    \end{aligned}
\end{flalign}
\end{center}
\end{problem}

\subsection{Markov Decision Process Formalization} \label{subsec:MDP}

We assume a Discrete Markov Decision Process (MDP) defined by a triplet $\{\mathcal{S}, \mathcal{A}, \mathcal{R}\}$: State space $\mathcal{S}$ is a set of states of the environment; Action space $\mathcal{A}$ is a set of actions from which the agent selects an action at each step; A reward function $\mathcal{R}$ that computes the reward of an action $a_i$ from state $s_i$ to $s'_i$, $R_i=r(s_i,a_i,s'_i )$. \\

\noindent{\bf States and actions.} We define an exploratory agent’s environment as a set of distinct quizzes, each containing $k$ MCQs. Although our model is not restricted to pre-existing quizzes, in our implementation, we materialize the space of all possible quizzes to achieve efficiency. The state space represents a quiz $Z$ as a set of $k$ MCQs, and each state is the concatenation of its quiz-level topic and difficulty distribution vectors. When an agent visits a state $s$ (i.e., a quiz $Z$), it seeks a better state $s'$ (a quiz $Z'$) by applying one of four actions: {\em SimTopic, SimLevel, DissTopic,} or {\em DissLevel}. Each action transforms a quiz $Z$ into a new quiz $Z'$ that is either similar or different with respect to topics or difficulty. Section~\ref{sec:setup} details how each action is efficiently implemented.\\

\noindent{\bf Reward design.} As {\sc QuizComp} is a multi-objective problem, we propose to define our reward using scalarization, a common approach that transforms the problem into a single objective via a weighted linear sum. Given a quiz $Z$, we can compute how close it is to the target coverage and difficulty:
\[\mathit{targetMatch}(Z, T_C, T_D) = \alpha \cdot \mathit{topicMatch}(Z,T_C) + 
\]
\[(1 - \alpha) \cdot \mathit{diffMatch}(Z,T_D)
\],

where $\alpha \in [0,1]$. 

We define the reward of taking an action $a$ at a state $s$:
\[
   \mathcal{R} \leftarrow  \mathit{targetMatch}(s'.Z, T_C, T_D) - \]
   \[\mathit{targetMatch}(s.Z, T_C, T_D)
\].

A state $s \in \mathcal{S}$ contains a quiz $Z$, and applying action (a) yields a new state (s') with quiz $Z'$. The reward captures the agent’s progression: if $s'.Z$ is farther from the target than $s.Z$, the reward is negative, penalizing the action and discouraging its future use in similar situations. If $s'.Z$ is closer to the target, the agent receives a positive reward and is encouraged to reuse action $a$.\\


\noindent{\bf Exploration session.} An agent learns to navigate in the environment. In each step $i$, a new quiz $Z_{i+1}$ is composed based on the previous one $Z_i$ by taking an action $a_i$. An exploration session $S$, starting at state $s_1$ (i.e., defines a quiz $Z_1$), of length $n$, is a sequence of exploration states and actions: $
S = [(s_1, a_1), \dots,(s_n, a_n)].
$

\noindent{\bf Reinforcement Learning.} 
Model-free RL~\cite{sutton2018reinforcement} addresses sequential optimization by having an agent interact with an environment and maximize cumulative reward. We use this framework for {\sc QuizComp}, where the agent composes the best quiz (Z) (the best state (s)) by maximizing search progression. RL includes four elements: policy, reward, value function, and environment.

A policy maps perceived states (quizzes) to actions, sometimes via simple functions and sometimes via search, as in {\sc QuizComp}. The reward function defines the task objective. Maximizing total reward drives policy updates. While rewards capture immediate benefit, the value function reflects long-term desirability. Here, the agent starts from a random quiz $Z_1$ and moves closer to target topics and difficulties $T_C$ and $T_D$ at each step. Although RL can include a model predicting next states and rewards, we focus on model-free methods.

{\bf Policy $\pi$.} A policy $\pi : \mathcal{S} \times \mathcal{A} \rightarrow [0, 1]$ of an RL agent maps the probability of taking action $a \in \mathcal{A}$ in state $s \in \mathcal{S}$, that is, $\pi(s, a) = Pr(a_t = a | s_t = s)$.  

We can rewrite the definition of a session as $S^{\pi} = [(s_1, \pi(s_1)), \\ \dots,(s_n, \pi(s_n))]
$. By replacing each state by its respective quiz, we rewrite the session $S$ as $S^{\pi} = [(Z_1, \pi(Z_1)), \dots,(Z_n, \pi(Z_n))].
$

{\bf Optimal Policy $\pi^*$.}
A policy $\pi^*$ is optimal if its expected cumulative reward is greater than or equal to the expected cumulative reward of all other policies $\pi$.  The optimal policy has an associated optimal state-value function and optimal Q-function: $
Q^*(s, a)\leftarrow \mathit{max}_{\pi} Q_{\pi}(s, a).$

\vspace{-0.23cm}
\begin{problem}[Revisited \textsc{QuizComp} Problem]
Given a session $S$, we define $\mathit{sessionMatch}(.)$ to measure the agent’s progress toward finding a quiz that matches the target distributions, discounted by $\gamma \in[0, 1]$:

\begin{flalign}
\mathit{sessionMatch}(S, T_C, T_D) = \\ \sum_{(s_i, a_i) \in S} \gamma^{i} \big[ \mathit{targetMatch}(s_{i+1}.Z, T_C, T_D) -  \\ \mathit{targetMatch}(s_i.Z, T_C, T_D) \big]
\end{flalign}

Hence, the problem is to find an optimal policy $$
\pi^* = argmax_{\pi} \text{ }\mathit{sessionMatch}(S^{\pi}, T_C, T_D).
$$
\end{problem}

\subsection{Reinforcement Learning Solutions}
 \label{sec:RL}
We explore three RL solutions to solve {\sc QuizComp}.

{\em Deep Q-Network (DQN)}~\cite{sutton2018reinforcement} extends Q-learning using deep networks to approximate Q-values, estimating expected cumulative reward for each $\langle s,a\rangle$ pair. It iteratively updates Q-values to balance exploration and exploitation with learning rate $\alpha$ and discount factor $\gamma$:
\begin{equation}\label{qlrn} Q(s, a) \leftarrow Q(s, a) + \alpha \left[ R + \gamma \max_{a'\in A} Q(s', a') - Q(s, a) \right]. \end{equation}

We adopt PER~\cite{DBLP:journals/corr/SchaulQAS15} to enrich training experience.

{\em SARSA~\cite{sutton2018reinforcement}} is 
an on-policy variant of Q-learning, SARSA updates Q-values using the action the agent actually takes:
\begin{equation}\label{sarsa} Q(s, a) \leftarrow Q(s, a) + \alpha \left[ R + \gamma Q(s', a') - Q(s, a) \right]. \end{equation}

{\em A2C/A3C~\cite{DBLP:conf/icml/MnihBMGLHSK16}} are actor–critic methods that combine policy gradients with value estimation. The actor updates action probabilities while the critic estimates advantages:
\begin{equation}
\mathit{Advantage}(s,a) \approx R(s,a,s')
 + \gamma V(s') - V(s).
\end{equation}
 
A3C extends A2C with parallel asynchronous workers, each interacting with its own environment copy to stabilize learning and update shared actor–critic networks.

\section{Experiments}
\label{sec:exps}

\begin{table*}[htpb]
    \centering
    \renewcommand{\arraystretch}{1.2}
    \begin{small}
    \begin{tabular}{|l|p{4cm}|l|p{3cm}|p{3cm}|}
        \hline
        \textbf{Dataset} & \textbf{Description} & \textbf{\#MCQs} & \textbf{\#Topics} & \textbf{\#Diff. levels} \\
        \hline
        {\sc MeD}
        & generated using MedGemma 9B~\footnote{For anonymity reasons, we omit citing the original paper. It will be added after acceptance.} based on medical documents from the UNESS \url{https://entrainement-ecn.uness.fr/} platform for training medical professionals.
        & 
        1500+
        & 
        111 topics from which we sampled 10 topics, e.g., Cardiology, Rhumatology.
        & 
        The difficulty level of each MCQ has a numerical scale within $[1,5]$, 1 being the easiest.
        \\
        \hline
        
        {\sc Math}~\cite{dadosipb/PW3OWY_2024}  
        & collected from MathE \url{https://mathe.ipb.pt/} an e-learning platform for enhancing mathematical skills in higher education and supporting autonomous MCQ-based learning.
        & 1900+  
        & 15 topics from which we sampled 10 topics, e.g., Linear Algebra, Probabilities. 
        & The difficulty level of each MCQ on a increasing scale within $[1,6]$.\\
        \hline
    \end{tabular}
    \caption{Description of real datasets.}
    \label{tab:real-data}
    \end{small}
\end{table*}

\subsection{Setup}
\label{sec:setup}

{\bf Datasets.} 
We use three {\sc Synthetic} and two real datasets ({\sc MeD} and {\sc Math}, Table~\ref{tab:real-data}). While {\sc Math} has human-crafted MCQs, we generate MCQs for {\sc MeD} using MedGemma 9B, performing comparably to GPT-4o. Synthetic datasets control data distributions (uniform or biased) to study agent behavior (Table~\ref{tab:datasets}), sampling topic and difficulty vectors via a multivariate Dirichlet (parameters 1 for uniform, 0.5 for biased). Our real datasets have median similarity 0.67 ({\sc Math}) and 0.69 ({\sc MeD}), indicating generally similar quizzes. High-similarity quizzes ($>0.80$) are far fewer for biased targets; e.g., in {\sc MeD}, 3860 quizzes match $T_{\mathit{uniform}}$ vs. 86 for $T_{\mathit{bias}}$.

\begin{table*}[h]
\begin{center}
\begin{small}
\begin{tabular}{|l|l|l|}
\hline
\textbf{Dataset} & \textbf{Description} & \textbf{Example quiz representation} \\ \hline
{\sc Uniform} & Uniformly drawn MCQs &
\begin{tabular}[c]{@{}l@{}}
$T_C$ = \textless{}0.1, 0.1, 0.1, 0.1, 0.1, 0.1, 0.1, 0.1, 0.1, 0.1\textgreater\\
$T_D$ = \textless{}0.2, 0.2, 0.2, 0.2, 0.2\textgreater
\end{tabular} \\ \hline
{\sc BiasedTopic} & Biased topics and uniform difficulties &
\begin{tabular}[c]{@{}l@{}}
$T_C$ = \textless{}0, 0, 0, 0, 0, 0.5, 0, 0, 0.5, 0\textgreater\\
$T_D$ = \textless{}0.2, 0.2, 0.2, 0.2, 0.2\textgreater
\end{tabular} \\ \hline
{\sc BiasedLevel} & Biased difficulties and uniform topics &
\begin{tabular}[c]{@{}l@{}}
$T_C$ = \textless{}0.1, 0.1, 0.1, 0.1, 0.1, 0.1, 0.1, 0.1, 0.1, 0.1\textgreater\\
$T_D$ = \textless{}0.5, 0, 0, 0, 0.5\textgreater
\end{tabular} \\ \hline
\end{tabular}
\caption{Description of {\sc Synthetic} datasets.}
    \label{tab:datasets}
    \end{small}
    \end{center}
\end{table*}

{\bf Teacher targets.}
We consider two types of target distributions. $T_{\mathit{uniform}}$ represents uniform coverage of all topics and difficulty levels, while $T_{\mathit{bias}}$ and $T'_{\mathit{bias}}$ represent biased targets, focusing on specific topics and difficulty levels.

{\bf Agents.}
We instantiate five different agents for each RL algorithm described in Section~\ref{sec:RL} by varying the weight of the reward $\alpha$ in $\{$0, 0.25, 0.5, 0.75, 1$\}$. Each instance of an algorithm focuses on giving more or less importance to optimizing one of the objectives based on its specific value of $\alpha$ (e.g., an instance with $\alpha=0$ optimizes only for the difficulty level and ignores topic distribution).

{\bf Implementation of environment and actions.}
We build the environment from MCQs by filtering those covering a random subset of $k=10$ topics yielding a universe to $10^4$ quizzes.  MDP actions use Cosine similarity to modify a quiz $Z$ into $Z'$, sampling the 25 most similar or dissimilar quizzes from the pre-sampled quizzes, while avoiding similarity greater than $0.95$. 

{\bf Training setup.}
We set the number of episodes to 5000 and the maximum number of iterations per episode to 100. We set the reward target threshold $\beta$ to $0.85$. This means that if the agent finds a quiz $Z$ with a similarity to the target that is equal to or higher than $0.85$, it considers it as a good result and stops the navigation. We used a $\epsilon$-greedy decay strategy for exploitation/exploration with $\epsilon\_decay=0.995$, $\epsilon\_min=0.05$. We set the value of $\gamma$ to $0.95$, the learning rate to $0.005$, and the batch size to $128$.

{\bf Oracle.}
We assume access to an oracle that always finds the closest quiz to the target, used to assess our agents’ ability to reach the ground truth. The oracle scans all 10,000 quizzes, but in real scenarios, the full quiz set may be unavailable, as quizzes are generated by the agent. Thus, the oracle serves only as an accuracy baseline.

{\bf Measures.}
We report measures for training: (1) evolution of Q-values across episodes to verify convergence, (2) action distribution, and for inference: (3) similarity to the best quiz, (4) \#iterations to reach it, and (5) avg time. All results are an aggregation over $10$ runs. 

\subsection{Summary of results}
Our results show that agents converge regardless of $\alpha$ values or dataset bias. The learned action distribution depends on $\alpha$ and the target but not the dataset: for $T_{\mathit{uniform}}$, agents favor similarity actions ({\em SimTopic}, {\em SimLevel}) and avoid dissimilarity actions, staying risk-averse, while for $T_{\mathit{bias}}$, they explore more using dissimilarity actions. In synthetic inference, RL agents find good quizzes independently of $\alpha$ or data bias, though matching topics is harder than difficulties due to their larger number. Real-world results confirm these findings: agents achieve good quizzes 100× faster than Oracle and reach high quiz similarity ($.925$) when increasing the reward threshold, likely due to replay buffers. Transfer learning is effective across datasets and from $T_{\mathit{bias}}$ to $T_{\mathit{uniform}}$, and between biased targets ($T_{\mathit{bias}}$ and $T'_{\mathit{bias}}$), though the reverse transfer is harder.

\subsection{Synthetic environments}
We report the DQN evaluation results with different values of $\alpha$ and different datasets. The other RL algorithms exhibit similar trends. The training is performed on $T_{\mathit{uniform}}$ as it reflects real settings. In the following, we report empirical results on both the training and inference phases.\newline

\begin{figure*}[htpb]
    \centering
    \includegraphics[width=0.65\linewidth]{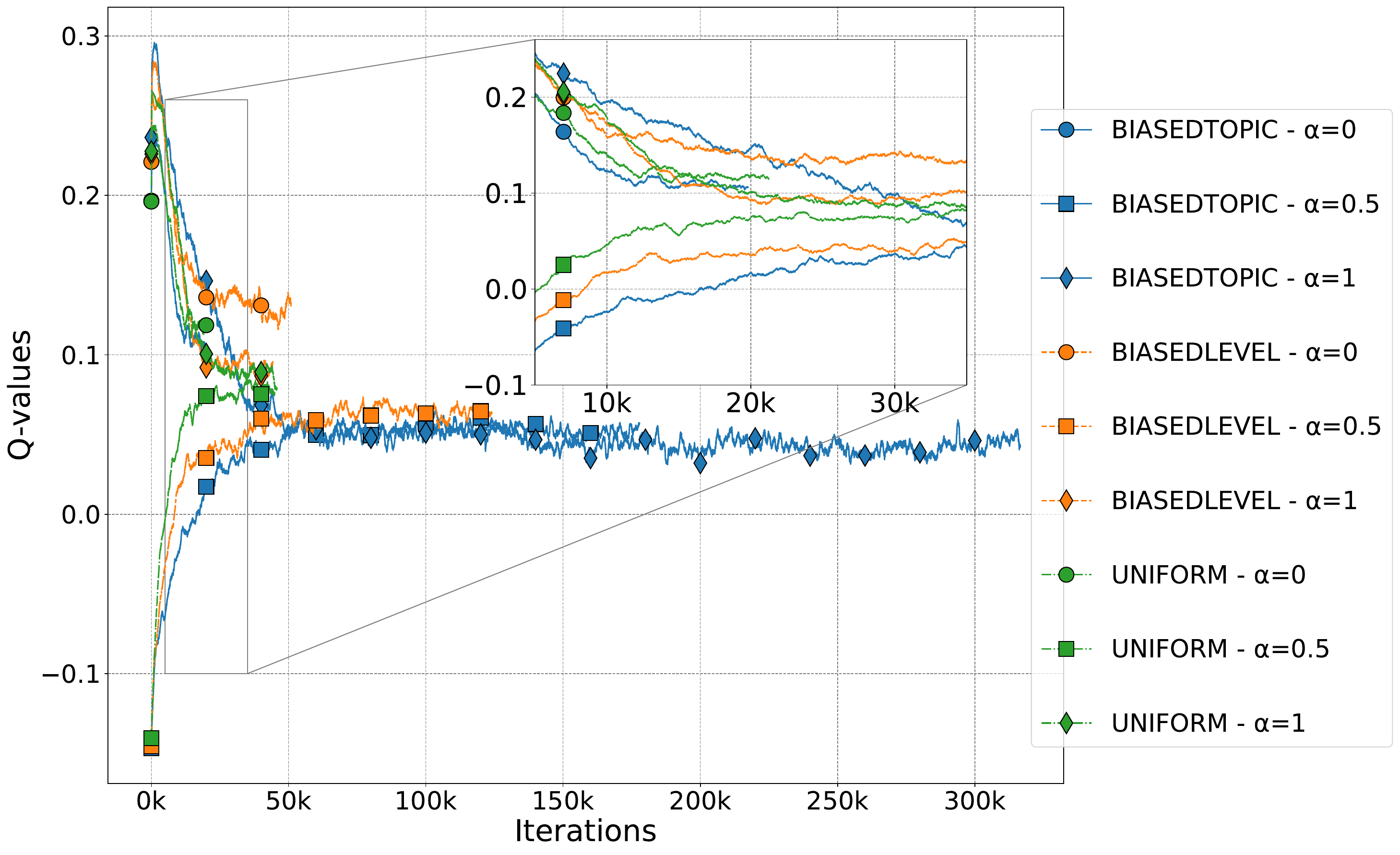}
    \caption{Evolution of max $Q$-function across training iterations of DQN on target $T_{\mathit{uniform}}$ for different datasets and $\alpha$ values.}
    \label{fig:train_qvalues}
\end{figure*}

\noindent{\it Training Phase.}
Figure~\ref{fig:train_qvalues} shows that all DQN agents converge around $\sim30$k steps across datasets and $\alpha$ values, despite early instability for $\alpha=0.5$ and biases like those in {\sc BiasedTopic}. Maximum Q-values vary with data bias and $\alpha$, but convergence remains consistent, highlighting agent stability without theoretical guarantees~\cite{dqn2013deepmind}.

\begin{figure*}[htpb]
    \centering
    \includegraphics[width=0.65\linewidth]{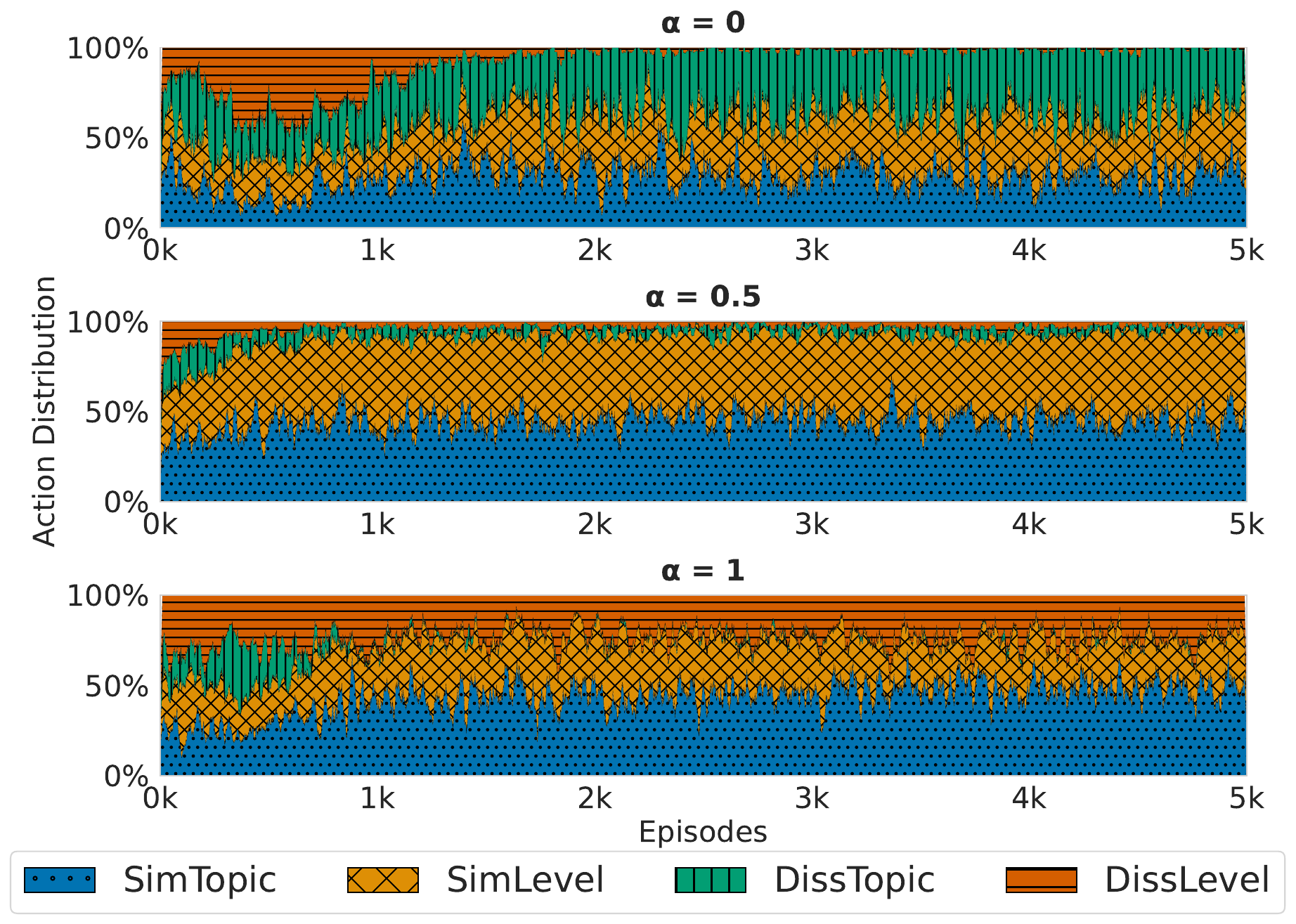}
    \caption{Evolution of learned actions per episode during DQN training on target $T_{\mathit{uniform}}$ for {\sc Uniform} dataset.}
    \label{fig:train_actions}
\end{figure*}

\begin{figure*}[htpb]
    \centering
    \includegraphics[width=0.65\linewidth]{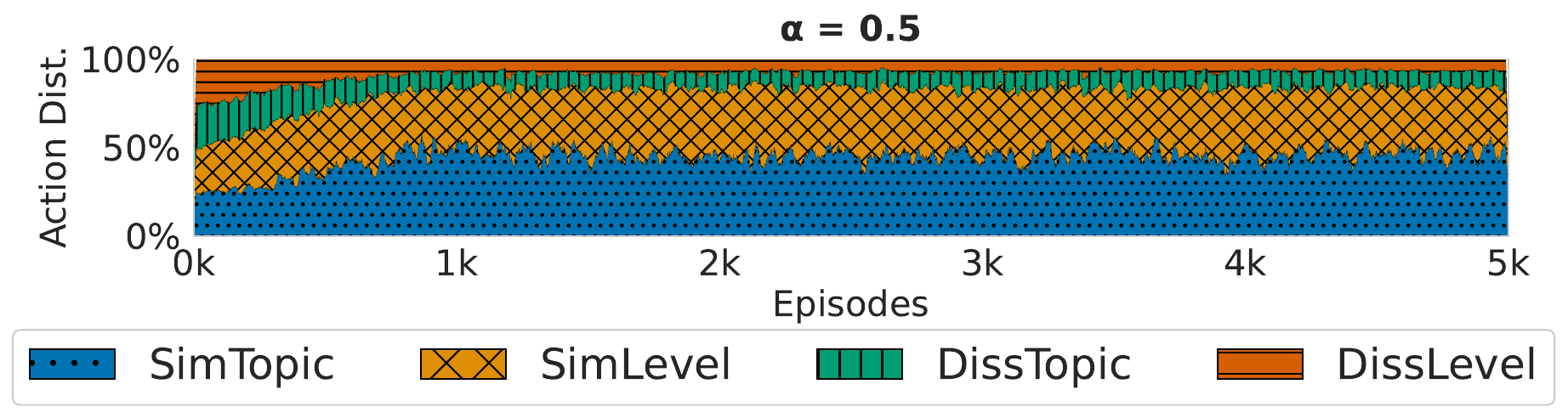}
    \caption{Evolution of learned actions per episode during DQN training on target $T_{\mathit{bias}}$ for {\sc Uniform} dataset.}
    \label{fig:train_actions_t4}
\end{figure*}

Figure~\ref{fig:train_actions} shows that for target $T_{\mathit{uniform}}$, agents primarily select similarity actions ({\em SimTopic}, {\em SimLevel}) and avoid dissimilarity actions ({\em DissTopic}, {\em DissLevel}) due to negative rewards, a pattern consistent across datasets and driven by $\alpha$ rather than data bias. For biased targets $T_{\mathit{bias}}$ (Figure~\ref{fig:train_actions_t4}), agents use more dissimilarity actions, reflecting exploratory behavior. Overall, learned policies depend on both the target type and the optimized objectives.
\\

\begin{table}[h]
\begin{small}
\begin{tabular}{|ll|l|l|l|}
\hline
\multicolumn{1}{|l|}{\bf Algos} & \multicolumn{1}{c|}{$\alpha$} & {\sc Uniform} & {\sc BiasedTopic} & {\sc BiasedLevel} \\ \hline
\multicolumn{1}{|c|}{\multirow{5}{*}{DQN}} & 0 & 0.896 & 0.903 & 0.888 \\ \cline{2-5} 
\multicolumn{1}{|c|}{} & 0.25 & 0.877 & 0.880 & 0.885 \\ \cline{2-5} 
\multicolumn{1}{|c|}{} & 0.5 & 0.870 & 0.870 & 0.862 \\ \cline{2-5} 
\multicolumn{1}{|c|}{} & 0.75 & 0.868 & 0.860 & 0.860 \\ \cline{2-5} 
\multicolumn{1}{|c|}{} & 1 & 0.877 & 0.851 & 0.869 \\ \hline \hline
\multicolumn{1}{|c|}{\multirow{5}{*}{Oracle}} & 0 & 0.998 & 0.997 & 1.000 \\ \cline{2-5} 
\multicolumn{1}{|c|}{} & 0.25 & 0.957 & 0.995 & 0.983 \\ \cline{2-5} 
\multicolumn{1}{|c|}{} & 0.5 & 0.944 & 0.994 & 0.975 \\ \cline{2-5} 
\multicolumn{1}{|c|}{} & 0.75 & 0.944 & 0.994 & 0.967 \\ \cline{2-5} 
\multicolumn{1}{|c|}{} & 1 & 0.973 & 0.999 & 0.974 \\ \hline
\end{tabular}
\caption{Average similarity with target $T_{\mathit{uniform}}$ for all {\sc Synthetic} datasets.}
\label{table:bias_infer_sim}
\end{small}
\end{table}

\begin{figure*}[htpb]
    \centering
    \includegraphics[width=0.8\linewidth]{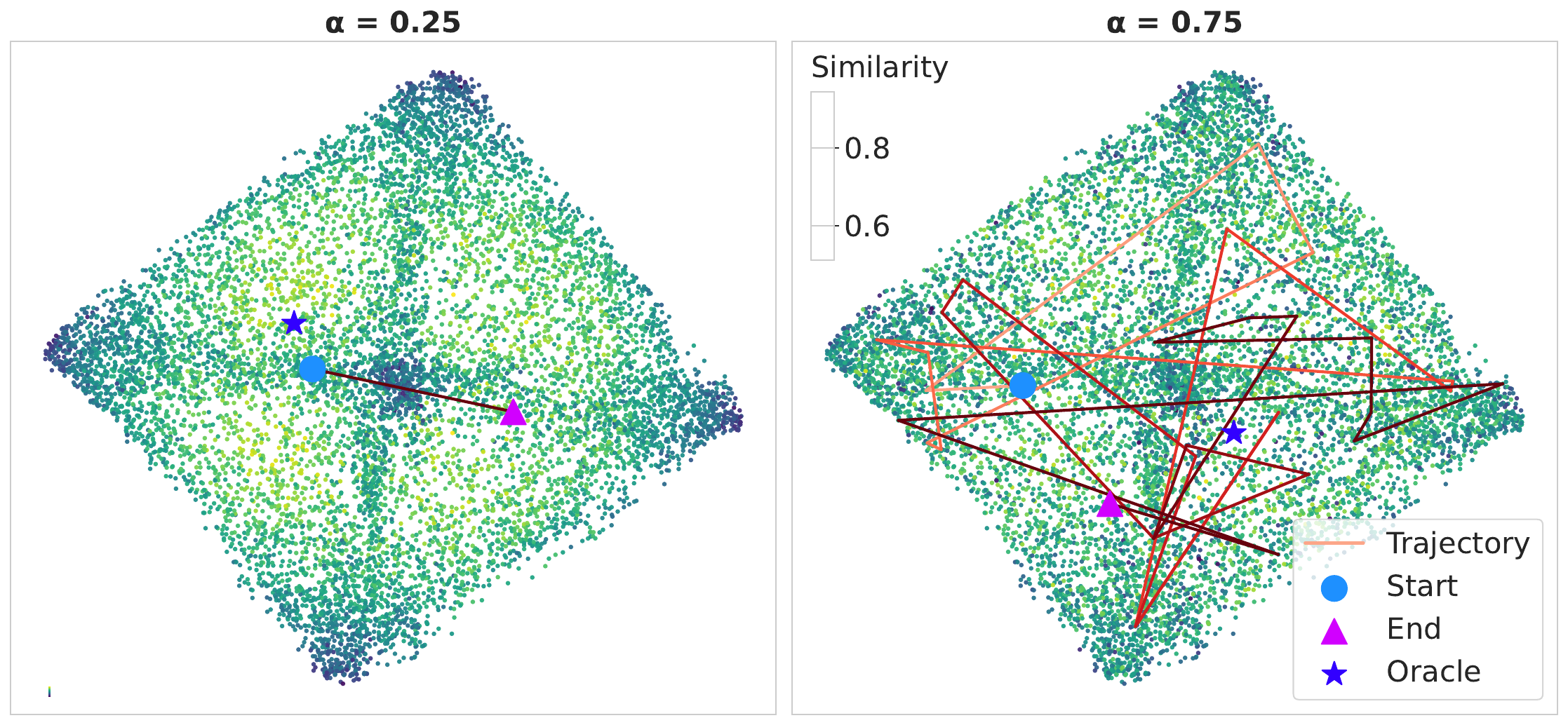}
    \caption{Examples of agent trajectories in {\sc Uniform} dataset using UMAP 2D projection.}
    \label{fig:agent_exploration}
\end{figure*}

\noindent{\it Inference Phase.} 
Table~\ref{table:bias_infer_sim} shows the average similarity of the found quizzes to the target $T_{\mathit{uniform}}$ using different $\alpha$ values on all {\sc Synthetic} datasets. Overall, the trained agents succeed in finding good quizzes close to the target $T_{\mathit{uniform}}$ as they reach the $0.85$ similarity threshold. The agents seem to find it more challenging to optimize for topics than for difficulty levels, as the similarity with $T_{\mathit{uniform}}$ slightly decreases when $\alpha$ increases (i.e., the increase observed with $\alpha=1$ is due to the fact that the agent optimizes only topic coverage). This can also be seen in Figure~\ref{fig:agent_exploration}, which depicts the inference trajectory that an agent takes to find $T_{\mathit{uniform}}$. The agent tends to use more actions with $\alpha=0.75$ than with $\alpha=0.25$. We explain this by the fact that there are more topics than difficulties, which makes it more challenging to match the desired topic distribution.

\subsection{Real environments}

\begin{table}[htpb]
\begin{small}
\begin{tabular}{|l|l|ccccc|}
\hline
\multirow{2}{*}{\bf Data} & \multirow{2}{*}{\bf Algos} & \multicolumn{5}{c|}{$\alpha$} \\ \cline{3-7} 
 &  & \multicolumn{1}{l|}{0} & \multicolumn{1}{l|}{0.25} & \multicolumn{1}{l|}{0.5} & \multicolumn{1}{l|}{0.75} & 1 \\ \hline
\multirow{3}{*}{{\sc MeD}} & DQN & \multicolumn{1}{l|}{\bf 0.918} & \multicolumn{1}{l|}{0.878} & \multicolumn{1}{l|}{0.871} & \multicolumn{1}{l|}{0.860} & \multicolumn{1}{l|}{0.893} \\ \cline{2-7} 
 & SARSA & \multicolumn{1}{l|}{0.911} & \multicolumn{1}{l|}{\bf 0.880} & \multicolumn{1}{l|}{\bf 0.876} & \multicolumn{1}{l|}{\bf 0.870} & 0.852 \\ \cline{2-7} 
 & A2C & \multicolumn{1}{l|}{0.907} & \multicolumn{1}{l|}{0.877} & \multicolumn{1}{l|}{0.869} & \multicolumn{1}{l|}{0.861} & 0.886\\ \cline{2-7} 
& A3C & \multicolumn{1}{l|}{0.910} & \multicolumn{1}{l|}{\bf 0.880} & \multicolumn{1}{l|}{0.869} & \multicolumn{1}{l|}{0.866} & {\bf 0.899} \\ \cline{2-7} 
 & Oracle & \multicolumn{1}{l|}{1.000} & \multicolumn{1}{l|}{0.978} & \multicolumn{1}{l|}{0.977} & \multicolumn{1}{l|}{0.988} & \multicolumn{1}{l|}{1.000} \\ \hline
 \hline
\multirow{3}{*}{{\sc Math}} & DQN & \multicolumn{1}{l|}{0.875} & \multicolumn{1}{l|}{\bf 0.859} & \multicolumn{1}{l|}{0.860} & \multicolumn{1}{l|}{0.858} & \multicolumn{1}{l|}{0.886} \\ \cline{2-7} 
 & SARSA & \multicolumn{1}{l|}{0.870} & \multicolumn{1}{l|}{0.856} & \multicolumn{1}{l|}{\bf 0.862} & \multicolumn{1}{l|}{0.852} & {0.906} \\ \cline{2-7}
  & A2C & \multicolumn{1}{l|}{0.870} & \multicolumn{1}{l|}{\bf 0.859} & \multicolumn{1}{l|}{0.855} & \multicolumn{1}{l|}{\bf 0.861} & 0.899\\ \cline{2-7} 
  
 & A3C & \multicolumn{1}{l|}{\bf 0.879} & \multicolumn{1}{l|}{ 0.856} & \multicolumn{1}{l|}{0.859} & \multicolumn{1}{l|}{0.852} & {\bf 0.913}  \\ \cline{2-7} 
 & Oracle & \multicolumn{1}{l|}{0.913} & \multicolumn{1}{l|}{0.913} & \multicolumn{1}{l|}{0.913} & \multicolumn{1}{l|}{0.936} & \multicolumn{1}{l|}{1.000} \\
 \hline
\end{tabular}
\caption{Average similarity with target for all agents in real datasets when finding $T_{\mathit{uniform}}$.}
\label{table:real_infer_sim}
\end{small}
\end{table}

Table~\ref{table:real_infer_sim} shows all algorithms achieve quiz similarity above 0.85. Differences between RL methods are small, showing they locate the target region effectively. RL finds good quizzes (e.g., $\alpha=0.5$, SARSA 0.876 vs. 0.977 best). Extreme $\alpha$ values improve similarity by focusing on a single objective, and no algorithm consistently outperforms others across datasets and $\alpha$ values.

\begin{table}[h!]
\begin{small}
\begin{tabular}{|l|l|l|l|}
\hline
{\bf Data} & {\bf Algos} & \multicolumn{1}{c|}{\# {\bf Iterations}} & \multicolumn{1}{c|}{\begin{tabular}[c]{@{}c@{}}{\bf Time} {\bf (sec)}\end{tabular}} \\ \hline
\multirow{3}{*}{\sc MeD} & DQN & 20 & 0.04 \\ \cline{2-4} 
 & SARSA & 29 & 0.07 \\ \cline{2-4} 
 & A2C & 25 & 0.05 \\ \cline{2-4} 
 & A3C & {\bf 17} & {\bf 0.04}\\ \cline{2-4}
 & Oracle & 10000 & 5.22\\ \hline \hline
\multirow{3}{*}{\sc Math} & DQN & 31 & 0.07 \\ \cline{2-4} 
 & SARSA & {\bf 30} & {\bf 0.06} \\ \cline{2-4} 
 & A2C & 33 & 0.06 \\ \cline{2-4}
 & A3C & 33 & 0.08\\ \cline{2-4}
  & Oracle & 10000 & 4.99 \\ \hline
\end{tabular}
\caption{Average \#iterations and time to find target $T_{\mathit{uniform}}$.}
\label{table:real_time}
\end{small}
\end{table}

Table~\ref{table:real_time} shows avg iterations and inference times for $T_{\mathit{uniform}}$. On {\sc Math}, algorithms perform similarly, while DQN and A3C slightly outperform on {\sc MeD}, likely due to DQN’s replay buffer~\cite{jimenez2023reinforcement} and A3C’s asynchronous exploration~\cite{DBLP:conf/icml/MnihBMGLHSK16}. Oracle finds the most similar quiz but scales poorly; e.g., on {\sc MeD}, DQN reaches 91.4\% of Oracle’s similarity about 100x faster.

\begin{figure}[htpb]
    \centering
    \includegraphics[width=\columnwidth]{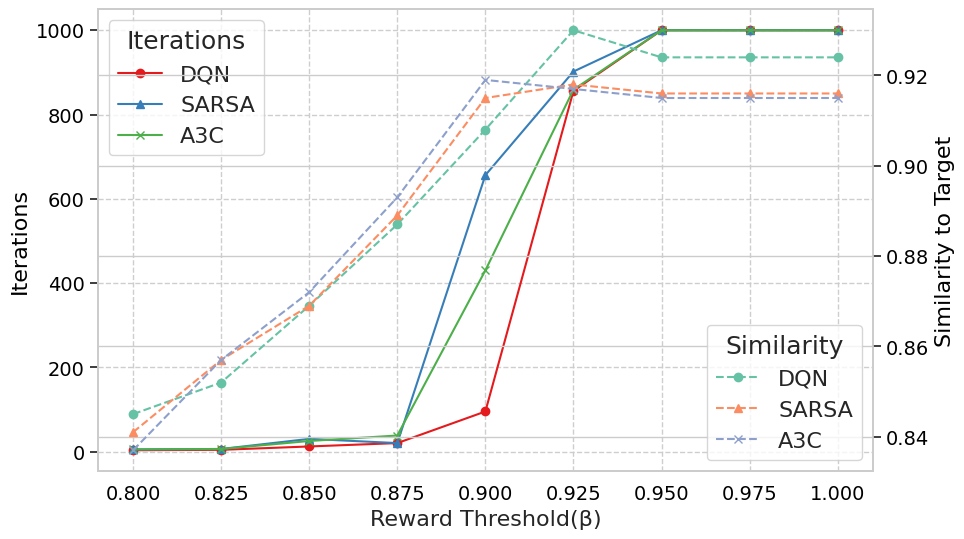}
    \caption{Evolution of number of iterations and similarity to target based on reward threshold for {\sc MeD} dataset.}
    \label{fig:thresholds}
\end{figure}

Small performance differences in Tables~\ref{table:real_infer_sim} may stem from the 0.85 reward threshold used in training and inference. Figure~\ref{fig:thresholds} shows that for thresholds below 0.925, DQN, SARSA, and A3C achieve similar similarity, while above 0.925, DQN outperforms due to stable learning with experience replay~\cite{jimenez2023reinforcement}; SARSA is conservative and A3C requires careful tuning~\cite{de2024comparative}. High-similarity quizzes ($>0.90$) often satisfy at least one criterion, limiting gains, and iteration counts rise with thresholds, justifying the 0.85 threshold for faster convergence and generalization.

\subsection{Transfer Learning}
We evaluate the transfer capabilities of a trained RL model across different targets ($T_{\mathit{uniform}}$,  $T_{\mathit{bias}}$, and $T'_{\mathit{bias}}$) and across datasets ({\sc MeD} and {\sc Math}). We set the value of $\alpha$ to $0.5$ to have an equi-balanced bi-objective optimization and rely only on real-world datasets. 

\begin{table}[htpb]
\begin{small}
\begin{tabular}{|l|l|l|l|l|}
\hline
\begin{tabular}[c]{@{}l@{}}{\bf Training}\\ {\bf target}\end{tabular} & \begin{tabular}[c]{@{}l@{}}{\bf Inference}\\ {\bf target}\end{tabular} & {\bf Algos} & \begin{tabular}[c]{@{}l@{}}{\bf Similarity}\\ {\bf w/target}\end{tabular} & {\bf \#Iter} \\ \hline
\multirow{3}{*}{$T_{\mathit{uniform}}$} & \multirow{3}{*}{$T_{\mathit{bias}}$} & DQN & 0.756 & 100 \\ \cline{3-5} 
 &  & SARSA & 0.791 & 100 \\ \cline{3-5} 
  &  & A2C & 0.800 & 83 \\ \cline{3-5} 
 &  & A3C & 0.775 & 95 \\ \cline{3-5}
 &  & Oracle & 0.888 & 10000\\ \hline \hline
\multirow{3}{*}{$T_{\mathit{bias}}$} & \multirow{3}{*}{$T_{\mathit{uniform}}$} & DQN & 0.866 & 24 \\ \cline{3-5} 
 &  & SARSA & 0.855 & 62 \\ \cline{3-5}
  &  & A2C & 0.852 & 67 \\ \cline{3-5}
 &  & A3C & 0.833 & 97 \\ \cline{3-5}
 &  & Oracle & 0.913 & 10000 \\ \hline \hline
  \multirow{3}{*}{$T_{\mathit{bias}}$} & \multirow{3}{*}{$T'_{\mathit{bias}}$} & DQN & 0.856 & 48 \\ \cline{3-5} 
 &  & SARSA & 0.851 & 84 \\ \cline{3-5} 
  &  & A2C & 0.866 & 52 \\ \cline{3-5} 
 &  & A3C & 0.866 & 50 \\ \cline{3-5}
 &  & Oracle & 0.927 & 10000 \\ \hline
\end{tabular}
\caption{Average similarity and $\#$Iter (iterations) for transferring the learning of agents across targets on {\sc Math} with $\alpha=0.5$.}
\label{table:transfer_tagerts}
\end{small}
\end{table}

Table~\ref{table:transfer_tagerts} shows transfer results on {\sc Math}. Transferring from a biased target to a uniform one works well, with agents reaching 0.85 similarity in few iterations. The reverse is harder: from $T_{\mathit{uniform}}$ to $T_{\mathit{bias}}$, agents either hit the iteration limit (DQN, SARSA) or need many iterations (A3C, 95). Biased targets require more exploration; agents trained on $T_{\mathit{uniform}}$ favor similarity actions and local search, while those trained on $T_{\mathit{bias}}$ use dissimilarity actions. Training on one biased target also transfers to another (e.g., $T'_{\mathit{bias}}$), with all agents reaching 0.85 similarity, showing a general policy for uniform or biased quizzes.

\begin{table}[htpb]
\begin{small}
\begin{tabular}{|l|l|l|l|l|}
\hline
\begin{tabular}[c]{@{}l@{}}{\bf Training}\\ {\bf data}\end{tabular} & \begin{tabular}[c]{@{}l@{}}{\bf Inference}\\ {\bf data}\end{tabular} & {\bf Algos} & \begin{tabular}[c]{@{}l@{}}{\bf Similarity}\\ {\bf w/target}\end{tabular} & {\bf \#Iter} \\ \hline
\multirow{3}{*}{\sc MeD} & \multirow{3}{*}{\sc Math} & DQN & 0.877 & 12 \\ \cline{3-5} 
 &  & SARSA & 0.881 & 8 \\ \cline{3-5} 
 &  & A2C & 0.865 & 16 \\ \cline{3-5} 
 &  & A3C & 0.864 & 11 \\ \cline{3-5} 
 &  & Oracle & 0.956 & 10000 \\ \hline \hline
\multirow{3}{*}{\sc Math} & \multirow{3}{*}{\sc MeD} & DQN & 0.868 & 20 \\ \cline{3-5} 
 &  & SARSA & 0.874 & 12 \\ \cline{3-5} 
   &  & A2C & 0.867 & 9 \\ \cline{3-5} 
 &  & A3C & 0.874 & 12 \\ \cline{3-5} 
 &  & Oracle & 0.977 & 10000 \\ \hline
\end{tabular}
\caption{Average similarity and $\#$Iter (Iterations) for transferring the learning of agents across datasets on $T_{\mathit{uniform}}$ with $\alpha=0.5$.}
\label{table:transfer_dataset}
\end{small}
\end{table}

Table~\ref{table:transfer_dataset} shows the results of transferring the learning of agents across datasets using $T_{\mathit{uniform}}$ as a target. In both cases  the similarity of the found quizzes is high even if the transfer is performed across two types of knowledge. This benefits from our formalization (Section~\ref{subsec:MDP}), which makes our agents agnostic to the types of MCQs or quizzes.

\subsection{Qualitative Study}
We launched a pilot user study with 28 participants to assess perceived workload using the NASA Task Load Index (NASA-TLX).\footnote{\url{https://en.wikipedia.org/wiki/NASA-TLX}} Participants were selected to have teaching experience at the university or secondary level and hold a Master's or PhD degree in mathematics or a closely related quantitative field.

After using our system to build quizzes, participants reported moderate mental demand and effort, low perceived controllability reflected in lower perceived performance and higher frustration, and low temporal demand, with moderate perceived accomplishment and slightly positive satisfaction.
Finally, 57\% of participants indicated they would use the system again, opening new opportunities for larger studies. 

\section{Related Work}
\noindent{\bf Standard quiz generation.} 
Early rule-based quiz generation  used templates and simple rules, such as “include two geometry questions,” offering automation but limited generalizability \cite{alsubait2016, Heilman2010}. Later, NLP and ML methods predicted question difficulty \cite{Feng2025} and used clustering or topic modeling to group or diversify questions \cite{Kurdi2020}. {\em These methods focus on single-objective optimization and do not support balancing multiple pedagogical criteria.}\\
\noindent{\bf Multi-objective quiz composition.} 
With a high-quality MCQ bank, focus shifts from generation to composition. MOEPG \cite{Shang2023} frames exam generation as multi-objective RL. {\em In our work, we target varied topic and difficulty distributions allowing us to train agents that capture different pedagogical goals. Additionally, we evaluate multiple RL algorithms.}\\
\noindent{\bf LLMs for quiz generation.} 
LLMs like GPT-4 generate MCQs text \cite{Meissner2024, Maity2024}, using prompt engineering or chain-of-thought \cite{Yao2024}, and can generate and evaluate quizzes \cite{Meissner2024, mucciaccia-etal-2025-automatic}, though irrelevant content may appear. Knowledge Tracing and RAG adapt quizzes to learners \cite{li2025tutorllmcustomizinglearningrecommendations}, while concept-based methods improve grounding \cite{fu2025conquerframeworkconceptbasedquiz}. {\em We use LLMs to generate MCQs but rely on RL for quiz composition to mimic a teacher balancing multiple objectives across data distributions.}

\section{Conclusion}
We investigated RL-based approaches for quiz composition and provided an extensive performance comparison of DQN, SARSA, and A2C/A3C. Our work investigated and demonstrated effectiveness of transfer learning across datasets and pedagogical targets. Our future work will broaden the action space and study trade-offs between on-the-fly MCQ generation with language models and meeting teacher objectives. We will compare the cost of prompt engineering and model calls to the cost of training an RL agent, including in transfer-learning settings, formalizing boundaries between training and reuse in line with recent work on ML reusability~\cite{DBLP:journals/vldb/Nikookar25}.


\section*{Impact Statement}
This paper advances Online RL by studying its application to teacher-facing services. Our work has notable societal implications, particularly in empowering teachers to understand their materials and generate personalized tests.


\newpage
\appendix
\onecolumn


\appendix

\section{Additional Synthetic Experiments and Qualitative Diagnostics}
\label{app:synthetic_extra}

This appendix complements the main experimental section with additional material focused
on the synthetic setting: (i) the distribution of the generated synthetic data, (ii) reward
stabilization during training, and (iii) qualitative visualizations of the agent navigation in the quiz space.

\subsection{Synthetic data generation and distributions}
\label{app:synthetic_generation}

\paragraph{Synthetic environments.}
We generate three synthetic datasets (\textsc{Uniform}, \textsc{BiasedTopic}, and
\textsc{BiasedLevel}) to study agent behavior under controlled data distributions.
We sample topic and difficulty vectors from a Dirichlet distribution, with concentration
parameters $\mu=1$ for \textsc{Uniform} data and $\mu=0.5$ for biased data.

\begin{figure*}[t]
    \centering
    \includegraphics[width=0.95\linewidth]{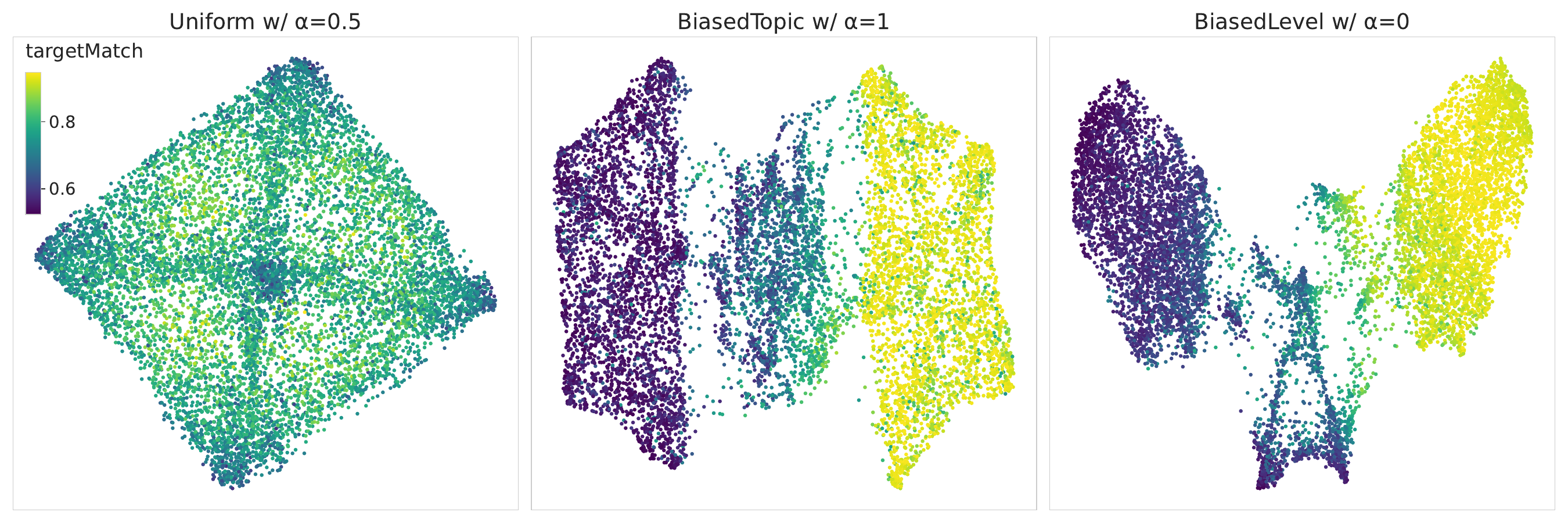}
    \caption{Projection of high-dimensional states in 2D with UMAP:
    a) {\sc Uniform} dataset with $\alpha=0.5$; b) {\sc BiasedTopic} dataset with $\alpha=1$;
    c) {\sc BiasedLevel} dataset with $\alpha=0$.}
    \label{fig:app_synthetic_datasets}
\end{figure*}

Table~\ref{tab:datasets} in the main paper reports the definition of the three synthetic datasets, while
Figure~\ref{fig:app_synthetic_datasets} provides a qualitative visualization of their
structure using a 2D UMAP projection.

\paragraph{Reward Definitions.}
We report results under two reward schemes. 
The first reward, denoted $R_1$, is a direct similarity reward baseline where the agent is rewarded according to the absolute similarity between the current quiz and the target distributions:
\begin{equation}
R_1(s,a,s') = \text{targetMatch}(Z', T_C, T_D),
\end{equation}
with
\begin{equation}
\text{targetMatch}(Z, T_C, T_D) = \alpha \cdot \text{topicMatch}(Z, T_C) + (1-\alpha)\cdot \text{diffMatch}(Z, T_D),
\end{equation}
where $\alpha \in [0,1]$ controls the relative importance of topic coverage versus difficulty alignment.

The second reward, denoted $R_2$, is a progress-based reward that captures the agent's improvement between consecutive quizzes:
\begin{equation}
R_2(s,a,s') = \text{targetMatch}(Z', T_C, T_D) - \text{targetMatch}(Z, T_C, T_D).
\end{equation}
Positive rewards indicate progress toward the target, while negative rewards penalize regressions. 
Unlike $R_1$, $R_2$ explicitly encourages exploratory trajectories by rewarding incremental improvements rather than absolute similarity.

\subsection{Reward stabilization during training}
\label{app:reward_stabilization}

\begin{figure*}[t]
    \centering
    \includegraphics[width=0.95\linewidth]{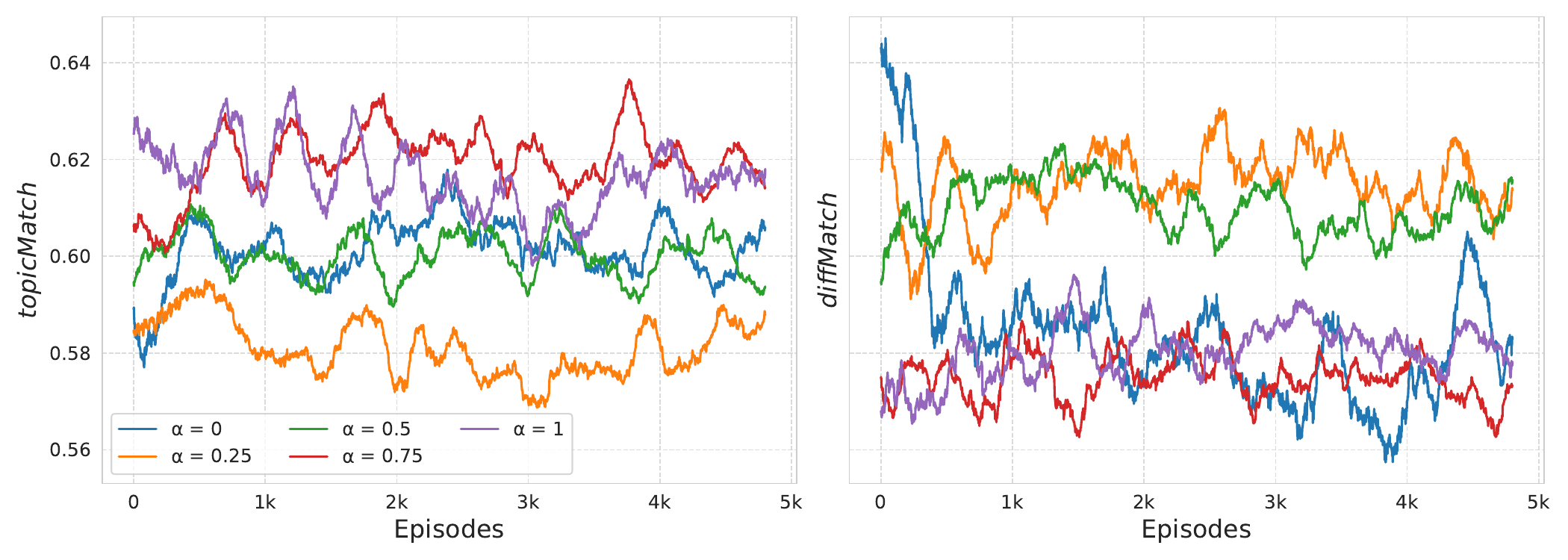}
    \caption{Evolution of \textit{topicMatch} and \textit{diffMatch} across training episodes of
    DQN on target $T_{\mathit{uniform}}$ for {\sc Uniform} dataset, with smoothing window of size 100.}
    \label{fig:app_reward_progress_r1}
\end{figure*}

Figure~\ref{fig:app_reward_progress_r1} shows the evolution of \textit{topicMatch} and
\textit{diffMatch} across training episodes under the direct reward baseline $\mathcal{R}_1$,
illustrating early stabilization and limited improvement.

\subsection{Agent trajectories in the quiz space}
\label{app:trajectories}

We provide qualitative visualizations of agent navigation both in the state space (UMAP)
and in the objective space (\textit{topicMatch}, \textit{diffMatch}).

\begin{figure*}[t]
    \centering
    \includegraphics[width=0.8\linewidth]{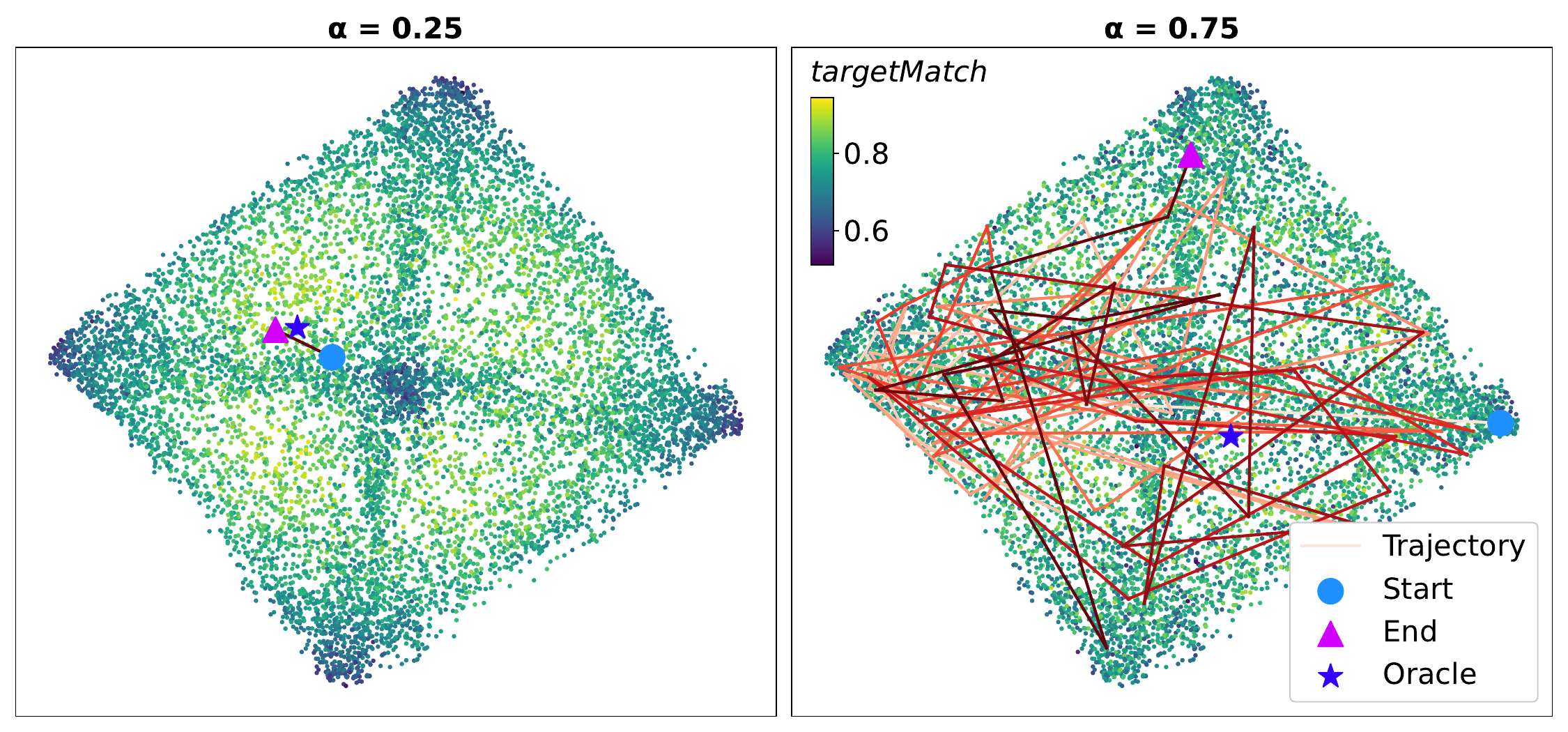}
    \caption{Examples of agent trajectories in {\sc Uniform} dataset using UMAP 2D projection following
    reward $\mathcal{R}_1$ (direct similarity reward); the colors of the agent trajectory represent the inference timeline, where
    time shifts from lighter to darker colors.}
    \label{fig:app_r1_agent_trajectory}
\end{figure*}

\begin{figure*}[t]
    \centering
    \includegraphics[width=0.8\linewidth]{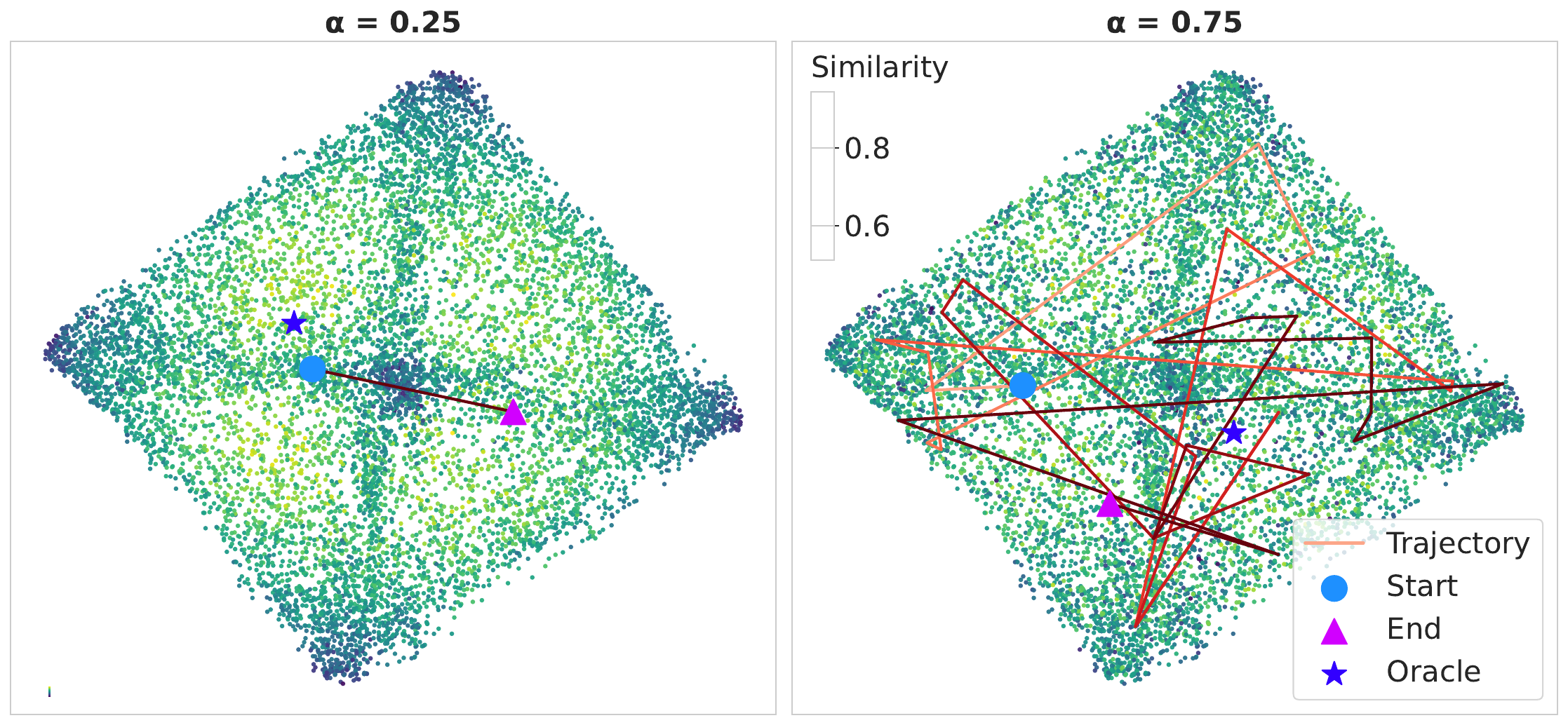}
    \caption{Examples of agent trajectories in {\sc Uniform} dataset using UMAP 2D projection following
    reward $\mathcal{R}_2$ (progress-based reward).}
    \label{fig:app_agent_exploration}
\end{figure*}

\paragraph{UMAP trajectories.}
Figures~\ref{fig:app_r1_agent_trajectory} and~\ref{fig:app_agent_exploration} show example
inference trajectories under $\mathcal{R}_1$ and $\mathcal{R}_2$, respectively.

\begin{figure*}[t]
    \centering
    \includegraphics[width=0.8\linewidth]{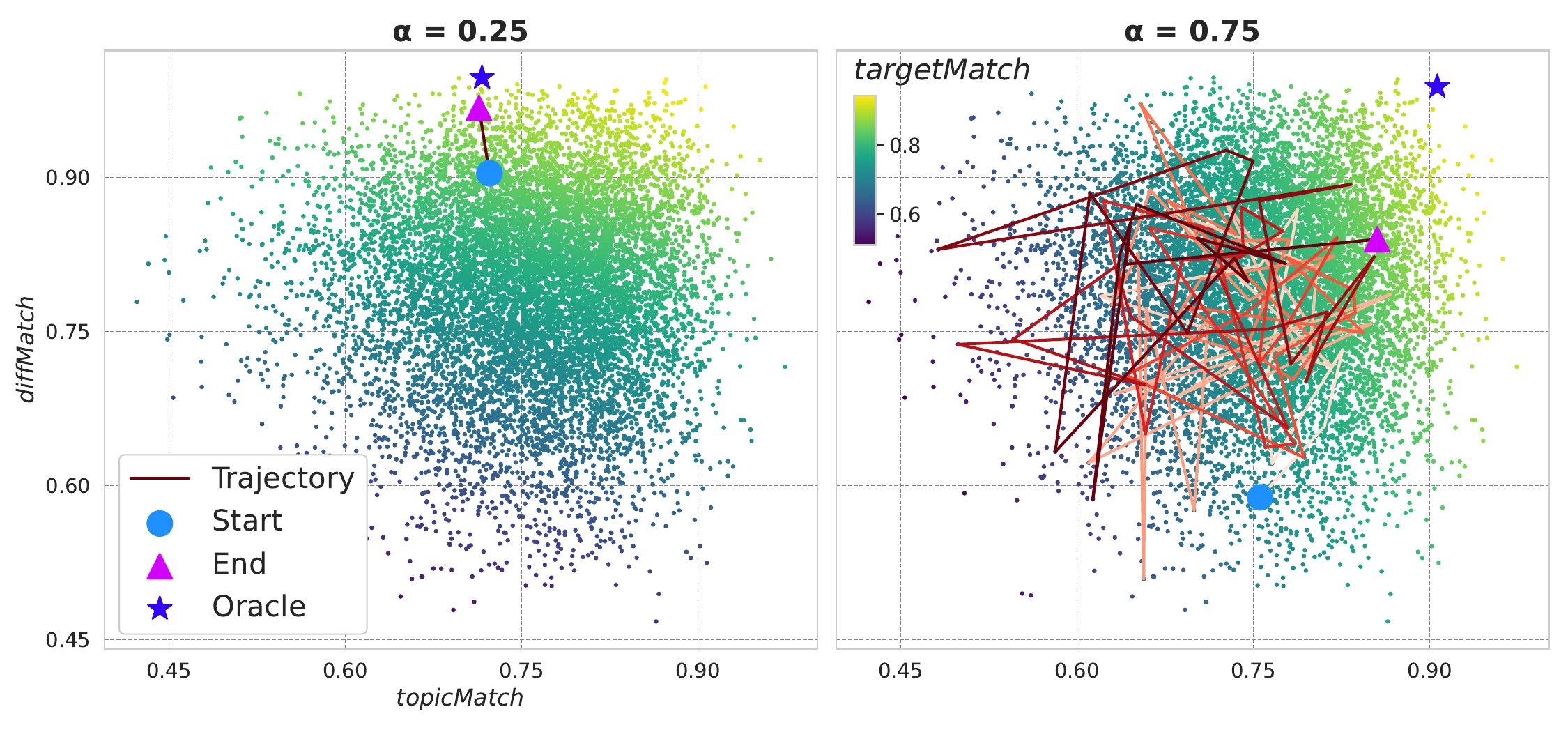}
    \caption{Examples of agent trajectories in {\sc Uniform} dataset using 2D projection with
    \textit{topicMatch} and \textit{diffMatch} following reward $\mathcal{R}_1$ (direct similarity reward).}
    \label{fig:app_r1_reward_landscape}
\end{figure*}

\begin{figure*}[t]
    \centering
    \includegraphics[width=0.8\linewidth]{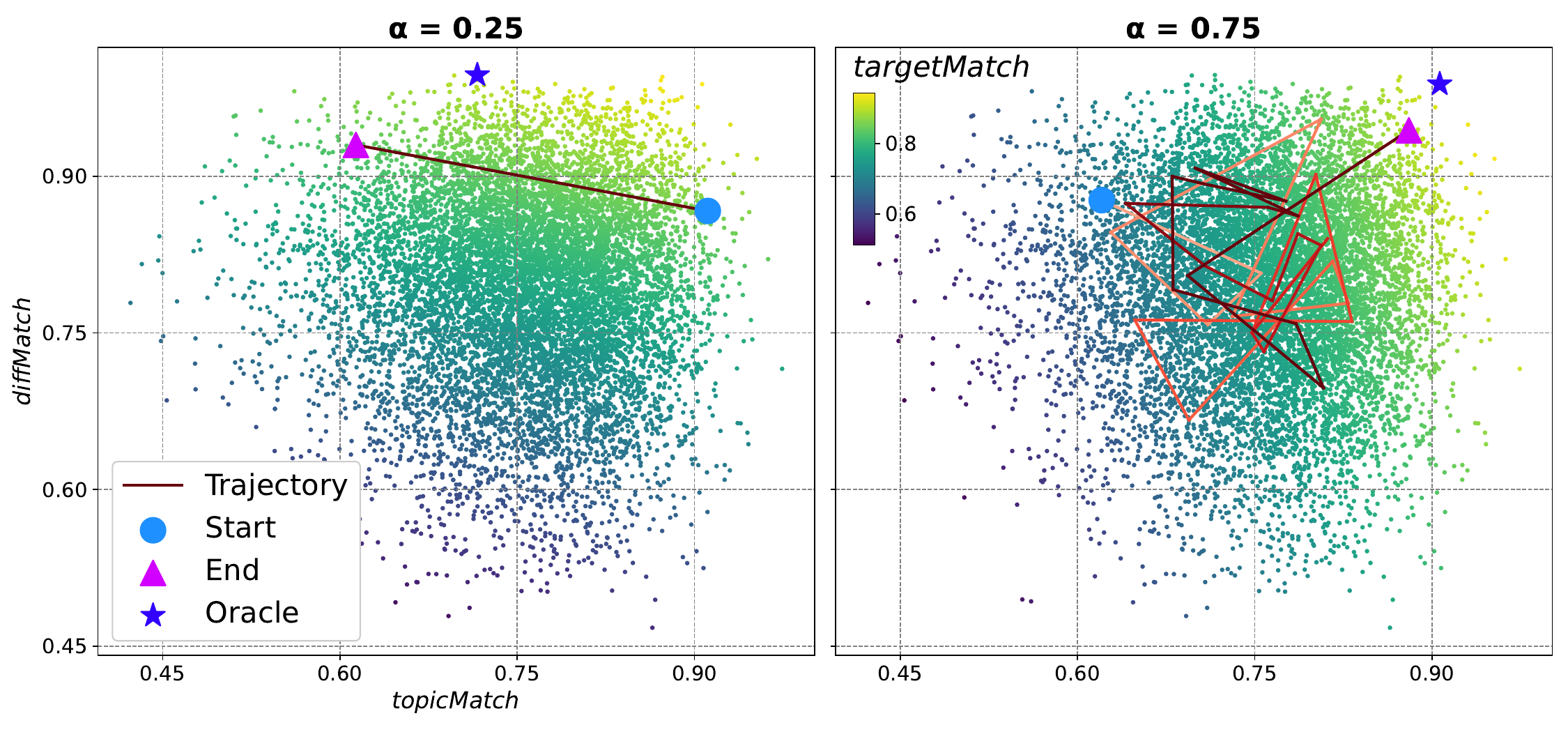}
    \caption{Examples of agent trajectories in {\sc Uniform} dataset using 2D projection with
    \textit{topicMatch} and \textit{diffMatch} following reward $\mathcal{R}_2$ (progress-based reward).}
    \label{fig:app_r2_reward_landscape}
\end{figure*}

\paragraph{Trajectories in objective space.}
Figures~\ref{fig:app_r1_reward_landscape} and~\ref{fig:app_r2_reward_landscape} show the same type
of analysis projected in the 2D plane defined by (\textit{topicMatch}, \textit{diffMatch}).

\end{document}